\documentclass[11pt,a4paper]{article}
\usepackage[hyperref]{emnlp2020}
\usepackage{times}
\usepackage{latexsym}

\usepackage{soul}
\usepackage{url}
\usepackage[utf8]{inputenc}
\usepackage{graphicx}
\usepackage{amsmath}
\usepackage{amsthm}
\usepackage{multirow}
\usepackage{booktabs}
\usepackage{amssymb}
\usepackage{tablefootnote}[2014/01/26]
\usepackage{algorithm}
\usepackage{algorithmic}
\usepackage{bm}
\usepackage[normalem]{ulem}
\newcommand{\Blue}[1]{\textcolor[rgb]{0.00,0.00,1.00}{#1}}

\newcommand{\comments}[1]{}

\newcommand{\yyc}[1]{\Blue{#1}}

\usepackage{microtype}

\aclfinalcopy 

\title{TinyBERT: Distilling BERT for Natural Language Understanding}

\author{Xiaoqi Jiao$^{1}$\thanks{\hspace{0.3pt} Authors contribute equally.} \hspace{0.3pt}\thanks{\hspace{0.3pt} This work is done when Xiaoqi Jiao is an intern at Huawei Noah's Ark Lab.}, Yichun Yin$^{2 \hspace{0.3pt} *} \hspace{0.3pt} \thanks{\hspace{0.5pt} Corresponding authors.}$, Lifeng Shang$^{2  \ddag}$, Xin Jiang$^{2}$\\
\textbf{Xiao Chen$^{2}$, Linlin Li$^{3}$, Fang Wang$^{1  \ddag}$ and Qun Liu$^{2}$}\\
$^{1}$Key Laboratory of Information Storage System, Huazhong University of\\ Science and Technology, Wuhan National Laboratory for Optoelectronics\\
$^{2}$Huawei Noah's Ark Lab\\
$^{3}$Huawei Technologies Co., Ltd.\\
\texttt{\{jiaoxiaoqi,wangfang\}@hust.edu.cn}\\
\texttt{\{yinyichun,shang.lifeng,jiang.xin\}@huawei.com}\\
\texttt{\{chen.xiao2,lynn.lilinlin,qun.liu\}@huawei.com}
}

\date{}

\begin{document}
\maketitle

\begin{abstract}
Language model pre-training, such as BERT, has significantly improved the performances of many natural language processing tasks. However, pre-trained language models are usually computationally expensive, so it is difficult to efficiently execute them on resource-restricted devices. To accelerate inference and reduce model size while maintaining accuracy, we first propose a novel {\it Transformer distillation} method that is specially designed for knowledge distillation (KD) of the Transformer-based models. By leveraging this new KD method, the plenty of knowledge encoded in a large ``teacher'' BERT can be effectively transferred to a small ``student'' TinyBERT. Then, we introduce a new {\it two-stage learning} framework for TinyBERT, which performs Transformer distillation at both the pre-training and task-specific learning stages. This framework ensures that TinyBERT can capture the general-domain as well as the task-specific knowledge in BERT.

TinyBERT$_{4}$\footnote{The code and models are publicly available at \url{https://github.com/huawei-noah/Pretrained-Language-Model/tree/master/TinyBERT}} with 4 layers is empirically effective and achieves more than 96.8\% the performance of its teacher BERT$_{\rm BASE}$ on GLUE benchmark, while being {\bf 7.5x smaller} and {\bf 9.4x faster} on inference. TinyBERT$_{4}$ is also significantly better than 4-layer state-of-the-art baselines on BERT distillation, with only {\bf $\sim$28\%} parameters and {\bf $\sim$31\%} inference time of them. Moreover, TinyBERT$_{6}$ with 6 layers performs {\bf on-par with} its teacher BERT$_{\rm BASE}$.
\end{abstract} 

\section{Introduction}\label{sec:intro}
Pre-training language models then fine-tuning on downstream tasks has become a new paradigm for natural language processing~(NLP). Pre-trained language models~(PLMs), such as BERT~\cite{devlin2019bert}, XLNet~\cite{yang2019xlnet}, RoBERTa~\cite{liu2019roberta}, ALBERT~\cite{Lan2020ALBERT}, T5~\cite{raffel2019exploring} and ELECTRA~\cite{clark2020electra}, have achieved great success in many NLP tasks (e.g., the GLUE benchmark~\cite{wang2018glue} and the challenging multi-hop reasoning task~\cite{ding2019cognitive}). 
However, PLMs usually have a large number of parameters and take long inference time, which are difficult to be deployed on edge devices such as mobile phones. Recent studies~\cite{kovaleva2019revealing,michel2019sixteen,voita2019analyzing} demonstrate that there is redundancy in PLMs. Therefore, it is crucial and feasible to reduce the computational overhead and model storage of PLMs while retaining their performances. 

There have been many model compression techniques~\cite{han2015deep} proposed to accelerate deep model inference and reduce model size while maintaining accuracy. The most commonly used techniques include quantization~\cite{gong2014compressing}, weights pruning~\cite{han2015learning}, and knowledge distillation~(KD)~\cite{romero2014fitnets}. In this paper, we focus on knowledge distillation, an idea originated from~\citet{hinton2015distilling}, in a {\it teacher-student} framework. KD aims to transfer the knowledge embedded in a large teacher network to a small student network where the student network is trained to reproduce the behaviors of the teacher network. Based on the framework, we propose a novel distillation method specifically for the Transformer-based models~\cite{vaswani2017attention}, and use BERT as an example to investigate the method for large-scale PLMs. 

KD has been extensively studied in NLP~\cite{kim2016sequence,hu2018attention} as well as for pre-trained language models \cite{sanh2019distilbert,sun2019patient,sun2020mobilebert,wang2020minilm}. The {\it pre-training-then-fine-tuning} paradigm firstly pre-trains BERT on a large-scale unsupervised text corpus, then fine-tunes it on task-specific dataset, which greatly increases the difficulty of BERT distillation. Therefore, it is required to design an effective KD strategy for both training stages.

\begin{figure}
  \centering
  \includegraphics[scale=0.4]{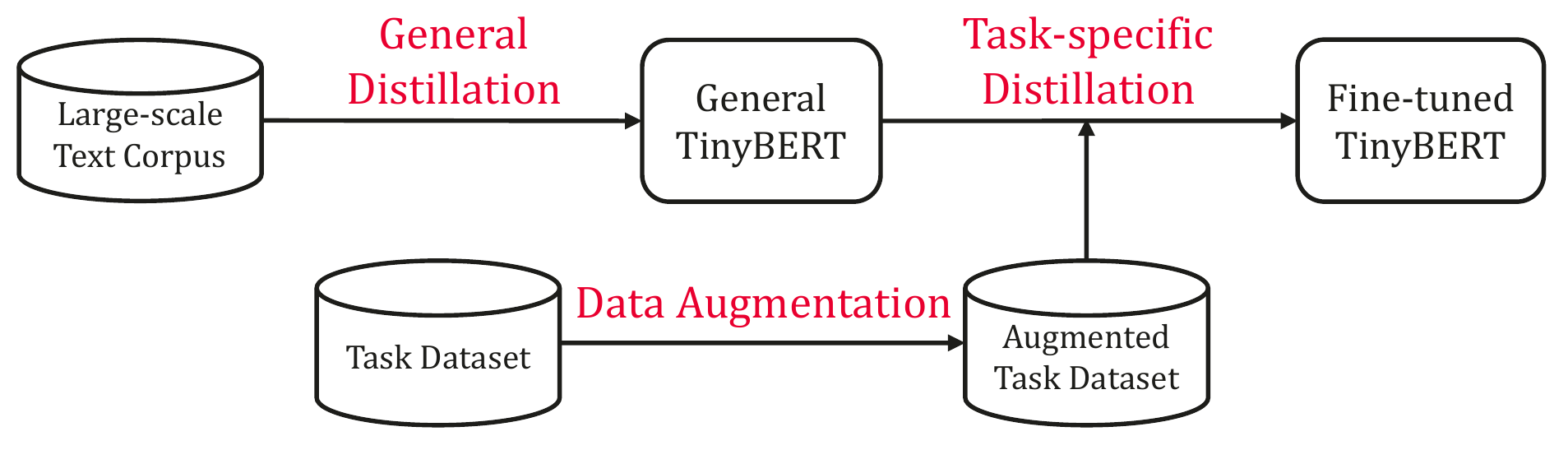}
  \caption{The illustration of TinyBERT learning.}
  \label{figure:tinybert_learning}
\end{figure}

To build a competitive TinyBERT, we firstly propose a new {\it Transformer distillation} method to distill the knowledge embedded in teacher BERT. Specifically, we design three types of loss functions to fit different representations from BERT layers: 1) the output of the embedding layer; 2) the hidden states and attention matrices derived from the Transformer layer; 3) the logits output by the prediction layer. The attention based fitting is inspired by the recent findings~\cite{clark2019does} that the attention weights learned by BERT can capture substantial linguistic knowledge, and it thus encourages the linguistic knowledge can be well transferred from teacher BERT to student TinyBERT. Then, we propose a novel {\it two-stage learning} framework including the {\it general distillation} and the {\it task-specific distillation}, as illustrated
in Figure~\ref{figure:tinybert_learning}. At general distillation stage, the original BERT without fine-tuning acts as the teacher model. The student TinyBERT mimics the teacher's behavior through the proposed Transformer distillation on general-domain corpus. After that, we obtain a general TinyBERT that is used as the initialization of student model for the further distillation. At the task-specific distillation stage, we first do the data augmentation, then perform the distillation on the augmented dataset using the fine-tuned BERT as the teacher model. It should be pointed out that both the two stages are essential to improve the performance and generalization capability of TinyBERT. 

%

The main contributions of this work are as follows: 1) We propose a new Transformer distillation method to encourage that the linguistic knowledge encoded in teacher BERT can be adequately transferred to TinyBERT; 2) We propose a novel two-stage learning framework with performing the proposed Transformer distillation at both the pre-training and fine-tuning stages, which ensures that TinyBERT can absorb both the general-domain and task-specific knowledge of the teacher BERT. 3) We show in the experiments that our TinyBERT$_{4}$ can achieve more than 96.8\% the performance of teacher BERT$_{\rm BASE}$ on GLUE tasks, while having much fewer parameters ($\sim$13.3\%) and less inference time ($\sim$10.6\%), and significantly outperforms other  state-of-the-art baselines with 4 layers on BERT distillation; 4) We also show that a 6-layer TinyBERT$_{6}$ can perform on-par with the teacher BERT$_{\rm BASE}$ on GLUE.

\section{Preliminaries}\label{sec:background}
In this section, we describe the formulation of Transformer~\cite{vaswani2017attention} and Knowledge Distillation~\cite{hinton2015distilling}. Our proposed {\it Transformer distillation} is a specially designed KD method for Transformer-based models. 

\subsection{Transformer Layer}
Most of the recent pre-trained language models (e.g., BERT, XLNet and RoBERTa) are built with Transformer layers, which can capture long-term dependencies between input tokens by self-attention mechanism. Specifically, a standard Transformer layer includes two main sub-layers: {\it multi-head attention}~(MHA) and {\it fully connected feed-forward} network~(FFN).

\noindent \textbf{Multi-Head Attention (MHA)}. The calculation of attention function depends on the three components of queries, keys and values, denoted as matrices $\bm{Q}$, $\bm{K}$ and $\bm{V}$ respectively. The attention function can be formulated as follows:
\begin{align}
	\label{eq:attention_score}
	\!\!\bm{A} \! & = \! \frac{\bm{Q}\bm{K}^{T}}{\sqrt{d_k}}, \\
	\!\!\texttt{Attention}(\bm{Q},\bm{K},\bm{V}) \! & = \! \texttt{softmax}(\bm{A})\bm{V},
\end{align}
where $d_k$ is the dimension of keys and acts as a scaling factor, $\bm{A}$ is the attention matrix calculated from the compatibility of $\bm{Q}$ and $\bm{K}$ by dot-product operation. The final function output is calculated as a weighted sum of values $\bm{V}$, and the weight is computed by applying {\tt softmax()} operation on the each column of matrix $\bm{A}$. According to Clark et al.~\shortcite{clark2019does}, the attention matrices in BERT can capture substantial linguistic knowledge, and thus play an essential role in our proposed distillation method.
 
Multi-head attention is defined by concatenating the attention heads from different representation subspaces as follows:
\begin{align}
\label{muti-head-attention}
\texttt{MHA}(\bm{Q},\bm{K},\bm{V}) \! &= \! \texttt{Concat}({\rm h}_1,\ldots,{\rm h}_k)\bm{W},
\end{align}
where $k$ is the number of attention heads, and ${\rm h}_i$ denotes the $i$-th attention head, which is calculated by the $\texttt{Attention}()$ function with inputs from different representation subspaces. The matrix $\bm{W}$ acts as a linear transformation.

\noindent \textbf{Position-wise Feed-Forward Network (FFN)}. Transformer layer also contains a fully connected feed-forward network, which is formulated as follows:
\begin{align}
\label{eq:FFN}
\texttt{FFN}(x) = \max(0, x\bm{W}_1 + b_1)\bm{W}_2 +b_2.
\end{align}
We can see that the FFN contains two linear transformations and one ReLU activation. 

\subsection{Knowledge Distillation}
KD aims to transfer the knowledge of a large teacher network $T$ to a small student network $S$. The student network is trained to mimic the behaviors of teacher networks. Let $f^{T}$ and $f^{S}$ represent the {\it behavior} functions of teacher and student networks, respectively. The behavior function targets at transforming network inputs to some informative representations, and it can be defined as the output of any layer in the network. In the context of Transformer distillation, the output of MHA layer or FFN layer, or some intermediate representations (such as the attention matrix $\bm{A}$) can be used as behavior function. Formally, KD can be modeled as minimizing the following objective function:
\begin{align}
\label{eq:point_distillation}
\mathcal{L}_{\text{KD}} =  \sum_{x \in \mathcal{X}} L\big(f^S(x), f^T(x)\big),
\end{align}
where $L(\cdot)$ is a loss function that evaluates the difference between teacher and student networks, $x$ is the text input and $\mathcal{X}$ denotes the training dataset. Thus the key research problem becomes how to define effective behavior functions and loss functions. Different from previous KD methods, we also need to consider how to perform KD at the pre-training stage of BERT in addition to the task-specific training stage.
\begin{figure}
  \centering
  \includegraphics[scale=0.38]{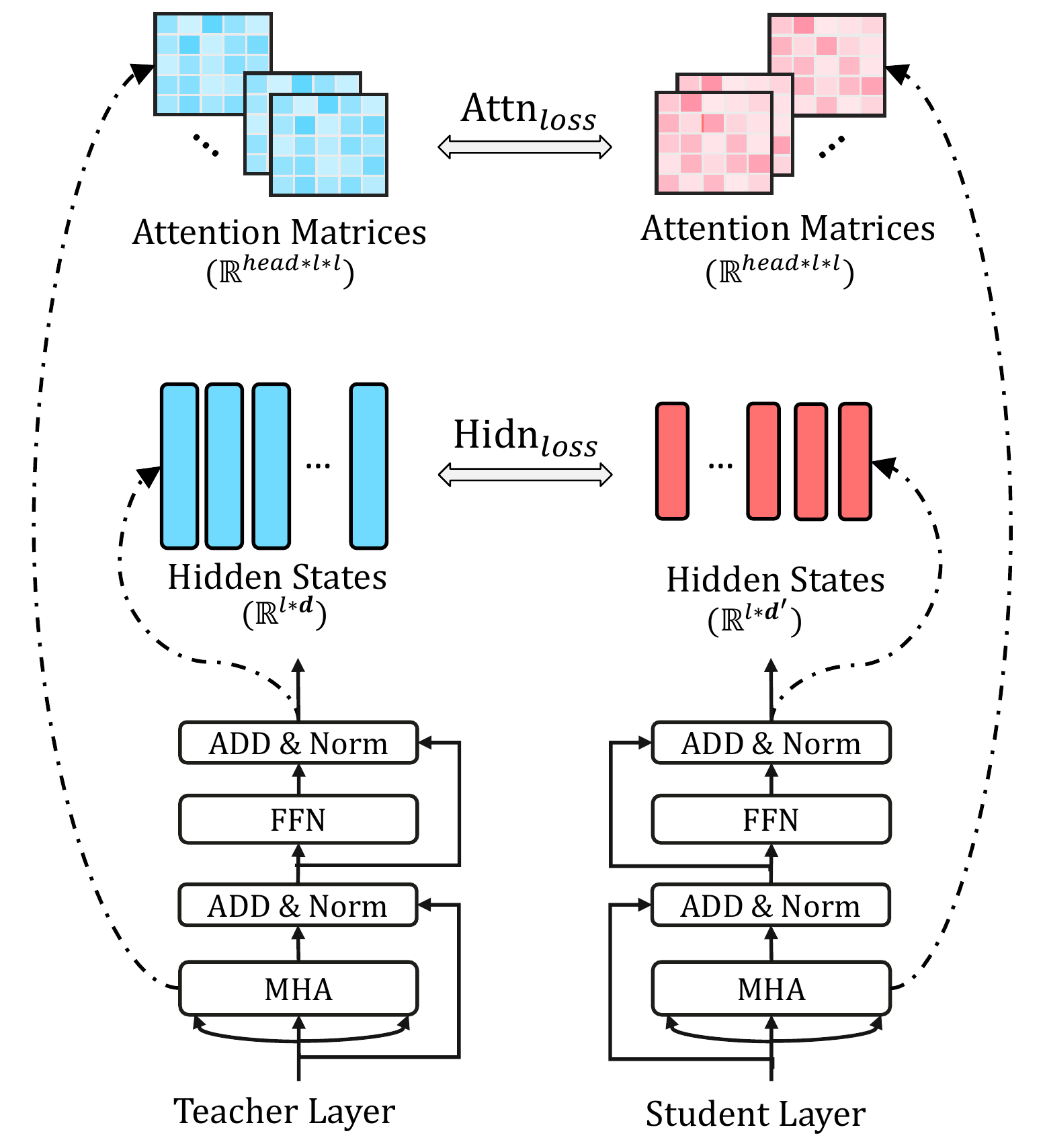}
  \caption{The details of Transformer-layer distillation consisting of Attn$_{loss}$({attention based distillation}) and Hidn$_{loss}$({hidden states based distillation}). }
  \label{figure:different_knowledge}
\end{figure}

\section{Method}\label{sec:method}
In this section, we propose a novel distillation method for Transformer-based models, and present a {\it two-stage learning} framework for our model distilled from BERT, which is called TinyBERT.

\subsection{Transformer Distillation}
The proposed {\it Transformer distillation} is a specially designed KD method for Transformer networks. In this work, both the student and teacher networks are built with Transformer layers. For a clear illustration, we formulate the problem before introducing our method.

\noindent{\bf Problem Formulation}. Assuming that the student model has $M$ Transformer layers and teacher model has $N$ Transformer layers, we start with choosing $M$ out of $N$ layers from the teacher model for the {\it Transformer-layer distillation}. Then a function $n=g(m)$ is defined as the mapping function between indices from student layers to teacher layers, which means that the $m$-th layer of student model learns the information from the $g(m)$-th layer of teacher model. To be precise, we set 0 to be the index of embedding layer and $M + 1$ to be the index of prediction layer, and the corresponding layer mappings are defined as $0=g(0)$ and $N+1=g(M+1)$ respectively. The effect of the choice of different mapping functions on the performances is studied in the experiment section. Formally, the student can acquire knowledge from the teacher by minimizing the following objective:
\begin{align}
\label{eq:kd_loss}
\!\!\!\mathcal{L}_{\text{model}} \! = \! \sum_{x \in \mathcal{X}} \sum^{M+1}_{m=0} \!\lambda_{m} \mathcal{L}_{\text{layer}}(f^S_m(x), f^T_{g(m)}(x)),
\end{align}
where $\mathcal{L}_{\text{layer}}$ refers to the loss function of a given model layer~(e.g., Transformer layer or embedding layer), $f_m(x)$ denotes the behavior function induced from the $m$-th layers and $\lambda_m$ is the hyper-parameter that represents the importance of the $m$-th layer's distillation.

\noindent{\bf Transformer-layer Distillation}. The proposed Transformer-layer distillation includes the {\it attention based distillation} and {\it hidden states based distillation}, which is shown in Figure~\ref{figure:different_knowledge}. The attention based distillation is motivated by the recent findings that attention weights learned by BERT can capture rich linguistic knowledge~\cite{clark2019does}. This kind of linguistic knowledge includes the syntax and coreference information, which is essential for natural language understanding. Thus we propose the attention based distillation to encourage that the linguistic knowledge can be transferred from teacher (BERT) to student (TinyBERT). Specifically, the student learns to fit the matrices of multi-head attention in the teacher network, and the objective is defined as:
\begin{align}
\label{eq:att_loss}
\mathcal{L}_{\text{attn}} = \frac{1}{h}\sum\nolimits^{h}_{i=1} \texttt{MSE}(\bm{A}_i^{S}, \bm{A}_i^{T}),
\end{align}
where $h$ is the number of attention heads, $\bm{A}_i \in \mathbb{R}^{l\times l} $ refers to the attention matrix corresponding to the $i$-th head of teacher or student, $l$ is the input text length, and {\tt MSE()} means the {\it mean squared error} loss function. In this work, the (unnormalized) attention matrix $\bm{A}_i$ is used as the fitting target instead of its softmax output $\texttt{softmax}(\bm{A}_i)$, since our experiments show that the former setting has a faster convergence rate and better performances. 

\comments{In this work, we use logits that is argument of the softmax function instead of softmax weight, as the fitting target. Because the experiments show that the logits fitting has faster convergence and better performances than the weight fitting. }

In addition to the attention based distillation, we also distill the knowledge from the output of Transformer layer, and the objective is as follows:
\begin{align}
\label{eq:hid_loss}
\mathcal{L}_{\text{hidn}} = \texttt{MSE}(\bm{H}^{S}\bm{W}_h, \bm{H}^{T}), 
\end{align}
where the matrices $\bm{H}^{S} \in \mathbb{R}^{l\times d'}$ and $\bm{H}^{T} \in \mathbb{R}^{l \times d}$ refer to the hidden states of student and teacher networks respectively, which are calculated by Equation~\ref{eq:FFN}. The scalar values $d$ and $d'$ denote the hidden sizes of teacher and student models, and $d'$ is often smaller than $d$ to obtain a smaller student network. The matrix $\bm{W}_h \in \mathbb{R}^{d' \times d} $ is a learnable linear transformation, which transforms the hidden states of student network into the same space as the teacher network's states. 

\noindent{\bf Embedding-layer Distillation}. Similar to the hidden states based distillation, we also perform embedding-layer distillation and the objective is:
\begin{align}
\label{eq:emb_loss}
\mathcal{L}_{\text{embd}} = \texttt{MSE}(\bm{E}^{S}\bm{W}_e, \bm{E}^{T}),
\end{align} where the matrices $\bm{E}^{S}$ and $\bm{H}^{T}$ refer to the embeddings of student and teacher networks, respectively. In this paper, they have the same shape as the hidden state matrices. The matrix $\bm{W}_e$ is a linear transformation playing a similar role as $\bm{W}_h$.
 
\comments{Following the teacher BERT's setting, we make the hidden size equal to the embedding size in student TinyBERT. Thus $\bm{E}^{S/T}$ have the same shapes with $\bm{H}^{S/T}$ respectively, and the learnable parameters $\bm{W}_e$ has the same shape with $\bm{W}_h$.} 

\noindent{\bf Prediction-layer Distillation}. In addition to imitating the behaviors of intermediate layers, we also use the knowledge distillation to fit the predictions of teacher model as in~\citet{hinton2015distilling}. Specifically, we penalize the soft cross-entropy loss between the student network's logits against the teacher's logits:
\begin{align}
\label{eq:pred_loss}
\mathcal{L}_{\text{pred}} = \texttt{CE}(\bm{z}^{T}/t, \bm{z}^{S}/t),
\end{align}
where ${\bm z}^{S}$ and ${\bm z}^{T}$ are the logits vectors predicted by the student and teacher respectively, \texttt{CE} means the \textit{cross entropy} loss, and $t$ means the temperature value. In our experiment, we find that $t=1$ performs well.

Using the above distillation objectives~(i.e. Equations~\ref{eq:att_loss}, \ref{eq:hid_loss}, \ref{eq:emb_loss} and \ref{eq:pred_loss}), we can unify the distillation loss of the corresponding layers between the teacher and the student network:
\begin{eqnarray}
\mathcal{L}_{\text{layer}} \!\! = \!\!
\begin{cases}
	\!\mathcal{L}_{\text{embd}},                                              	\!\!\!\!\!	& m \!= \!0              \\
	\!\mathcal{L}_{\text{hidn}} \! + \! \mathcal{L}_{\text{attn}},  			\!\!\!\!\!	& M \!\geq \!\!m \!> \!0 \\
	\!\mathcal{L}_{\text{pred}},                                    			\!\!\!\!\!	& m \!=\! M + 1
\end{cases}
\label{eq:model_loss}
\end{eqnarray}
\subsection{TinyBERT Learning}
The application of BERT usually consists of two learning stages: the pre-training and fine-tuning. The plenty of knowledge learned by BERT in the pre-training stage is of great importance and should be transferred to the compressed model. Therefore, we propose a novel two-stage learning framework including the {\it general distillation} and the {\it task-specific distillation}, as illustrated in Figure~\ref{figure:tinybert_learning}. General distillation helps TinyBERT learn the rich knowledge embedded in pre-trained BERT, which plays an important role in improving the generalization capability of TinyBERT. The task-specific distillation further teaches TinyBERT the knowledge from the fine-tuned BERT. With the two-step distillation, we can substantially reduce the gap between teacher and student models. 

{\bf General Distillation}. We use the original BERT without fine-tuning as the teacher and a large-scale text corpus as the training data. By performing the Transformer distillation~\footnote{In the general distillation, we do not perform prediction-layer distillation as Equation~\ref{eq:pred_loss}. Our  motivation is to make the TinyBERT primarily learn the intermediate structures of BERT at pre-training stage. From our preliminary experiments, we also found that conducting prediction-layer distillation at pre-training stage does not bring extra improvements on downstream tasks, when the Transformer-layer distillation (Attn and Hidn distillation) and Embedding-layer distillation have already been performed.} on the text from general domain, we obtain a general TinyBERT that can be fine-tuned for downstream tasks. However, due to the significant reductions of the hidden/embedding size and the layer number, general TinyBERT performs generally worse than BERT. 

{\bf Task-specific Distillation}. Previous studies show that the complex models, such as fine-tuned BERTs, suffer from over-parametrization for domain-specific tasks~\cite{kovaleva2019revealing}. Thus, it is possible for smaller models to achieve comparable performances to the BERTs. To this end, we propose to  produce competitive fine-tuned TinyBERTs through the task-specific distillation. In the task-specific distillation, we re-perform the proposed Transformer distillation on an augmented task-specific dataset. Specifically, the fine-tuned BERT is used as the teacher and a data augmentation method is proposed to expand the task-specific training set. Training with more task-related examples, the generalization ability of the student model can be further improved.

\begin{algorithm}[tb]
\algsetup{linenosize=\small} \small
\caption{Data Augmentation Procedure for Task-specific Distillation}
\label{alg:algorithm}
\textbf{Input}:  $\mathbf{x}$ is a sequence of words\\
\textbf{Params}:  $p_{t}$: the threshold probability\\ 
\hspace*{1.1cm}   $N_{a}$: the number of samples augmented per example\\ 
\hspace*{1.1cm}	  $K$: the size of candidate set \\
\textbf{Output}: ${D^{\prime}}$: the augmented data
\begin{algorithmic}[1] 
\STATE $n \gets 0\ ;\ \ D^{\prime} \gets [\ ]$
\WHILE{$n < N_{a}$}
\STATE $\mathbf{x}_{m} \gets \mathbf{x}$
\FOR{$i \gets $1\ to\ len$(\mathbf{x})$}
	\IF{$\mathbf{x}[i]$ \ is\ a\ single-piece\ word}
     	\STATE Replace $\mathbf{x}_{m}[i]$ with $\texttt{[MASK]}$
	\STATE $C \gets K$ most probable words of $\texttt{BERT}(\mathbf{x}_{m})[i]$
	\ELSE
		\STATE $C \gets K$ most similar words of $\mathbf{x}[i]$ from GloVe
	\ENDIF
	\STATE Sample $ p \sim$ Uniform(0, 1)
	\IF{$p \leq p_{t}$}
     		\STATE Replace $\mathbf{x}_{m}[i]$ with a word in $C$ randomly
	\ENDIF
\ENDFOR
\STATE Append $\mathbf{x}_{m}$ to ${D^{\prime}}$
     	\STATE $n \gets n + 1$
\ENDWHILE
\STATE \textbf{return} $D^{\prime}$
\end{algorithmic}
\end{algorithm}

{\bf Data Augmentation}. We combine a pre-trained language model BERT and GloVe~\cite{pennington2014glove} word embeddings to do word-level replacement for data augmentation. Specifically, we use the language model to predict word replacements for single-piece words~\cite{wu2019conditional}, and use the word embeddings to retrieve the most similar words as word replacements for multiple-pieces words\footnote{A word is tokenized into multiple word-pieces by the tokenizer of BERT.}. Some hyper-parameters are defined to control the replacement ratio of a sentence and the amount of augmented dataset. More details of the data augmentation procedure are shown in Algorithm~\ref{alg:algorithm}. We set $p_{t}$ = 0.4, $N_{a}$ = 20, $K$ = 15 for all our experiments.

The above two learning stages are complementary to each other: the general distillation provides a good initialization for the task-specific distillation, while the task-specific distillation on the augmented data further improves TinyBERT by focusing on learning the task-specific knowledge. Although there is a  significant reduction of model size, with the data augmentation and by performing the proposed Transformer distillation method at both the pre-training and fine-tuning stages, TinyBERT can achieve competitive performances in various NLP tasks.

\section{Experiments}\label{sec:exp}
\begin{table*}[tbp]
\begin{center}
\scalebox{0.72}{
\begin{tabular}{@{}l|ccc|cccccccc|c@{}}
	\textbf{System}                   & \textbf{\#Params} & \textbf{\#FLOPs} & \textbf{Speedup} &    \textbf{MNLI-(m/mm)}     & \textbf{QQP}  & \textbf{QNLI} & \textbf{SST-2} & \textbf{CoLA} & \textbf{STS-B} & \textbf{MRPC} & \textbf{RTE}  & \textbf{Avg}  \\ \hline
	BERT$_{\rm BASE}$ (Teacher)       &       109M        &      22.5B       &       1.0x       &          83.9/83.4          &     71.1      &     90.9      &      93.4      &     52.8      &      85.2      &     87.5      &     67.0      &     79.5      \\ \hline
	BERT$_{\rm TINY}$                &       14.5M       &       1.2B       &       9.4x       &          75.4/74.9          &     66.5      &     84.8      &      87.6      &     19.5      &      77.1      &     83.2      &     62.6      &     70.2      \\
	
	BERT$_{\rm SMALL}$                &       29.2M       &       3.4B       &       5.7x       &          77.6/77.0          &     68.1      &     86.4      &      89.7      &     27.8      &      77.0      &     83.4      &     61.8      &     72.1      \\
	
	BERT$_{4}$-PKD                    &       52.2M       &       7.6B       &       3.0x       &          79.9/79.3          &     70.2      &     85.1      &      89.4      &     24.8      &      79.8      &     82.6      &     62.3      &     72.6      \\
	DistilBERT$_{4}$                  &       52.2M       &       7.6B       &       3.0x       &          78.9/78.0          &     68.5      &     85.2      &      91.4      &     32.8      &      76.1      &     82.4      &     54.1      &     71.9      \\
	MobileBERT$_{\rm TINY} \dagger$   &       15.1M       &       3.1B       &        -         &          81.5/81.6          &     68.9      & \textbf{89.5} &      91.7      & \textbf{46.7} &      80.1      & \textbf{87.9} &     65.1      & \textbf{77.0} \\
	TinyBERT$_{4}$ (ours)             &       14.5M       &       1.2B       &       9.4x       & \textbf{82.5}/\textbf{81.8} & \textbf{71.3} &     87.7      & \textbf{92.6}  &     44.1      & \textbf{80.4}  &     86.4      & \textbf{66.6} & \textbf{77.0} \\ \hline
	BERT$_{6}$-PKD                    &       67.0M       &      11.3B       &       2.0x       &          81.5/81.0          &     70.7      &     89.0      &      92.0      &       -       &       -        &     85.0      &     65.5      &       -       \\
	PD                    &       67.0M       &      11.3B       &       2.0x       &          82.8/82.2          &     70.4      &     88.9      &       91.8      &       -       &       -        &     86.8      &     65.3      &       -       \\

	DistilBERT$_{6}$                  &       67.0M       &      11.3B       &       2.0x       &          82.6/81.3          &     70.1      &     88.9      &      92.5      &     49.0      &      81.3      &     86.9      &     58.4      &     76.8      \\
	TinyBERT$_{6}$ (ours)             &       67.0M       &      11.3B       &       2.0x       & \textbf{84.6}/\textbf{83.2} & \textbf{71.6} & \textbf{90.4} & \textbf{93.1}  & \textbf{51.1} & \textbf{83.7}  & \textbf{87.3} & \textbf{70.0} & \textbf{79.4}
\end{tabular}
}
\caption{Results are evaluated on the test set of GLUE official benchmark. The best results for each group of student models are in-bold. The architecture of TinyBERT$_{4}$ and BERT$_{\rm TINY}$ is ($M$=4, $d$=312, $d_i$=1200), BERT$_{\rm SMALL}$ is ($M$=4, $d$=512, $d_i$=2048), BERT$_{4}$-PKD and DistilBERT$_{4}$ is ($M$=4, $d$=768, $d_i$=3072) and the architecture of BERT$_{6}$-PKD, DistilBERT$_{6}$ and TinyBERT$_{6}$ is ($M$=6, $d$=768, $d_i$=3072). All models are learned in a single-task manner. The inference speedup is evaluated on a single NVIDIA K80 GPU.  $\dagger$~denotes that the comparison between MobileBERT$_{\rm TINY}$ and TinyBERT$_4$ may not be fair since the former has 24 layers and is task-agnosticly distilled from IB-BERT$_{\rm LARGE}$ while the later is a 4-layers model task-specifically distilled from BERT$_{\rm BASE}$.} 
\label{tab:glue_main}
\vspace{-0.1in}
\end{center}
\end{table*}

In this section, we evaluate the effectiveness and efficiency of TinyBERT on a variety of tasks with different model settings. 

\subsection{Datasets}\label{subsec:datasets}

We evaluate TinyBERT on the General Language Understanding Evaluation (GLUE)~\cite{wang2018glue} benchmark, which consists of 2 single-sentence tasks: CoLA~\cite{warstadt2019neural}, SST-2~\cite{socher2013recursive}, 3 sentence similarity tasks: MRPC~\cite{dolan2005automatically}, STS-B~\cite{cer2017semeval}, QQP~\cite{chen2018quora}, and 4 natural language inference tasks: MNLI~\cite{williams2018broad}, QNLI~\cite{rajpurkar2016squad}, RTE~\cite{bentivogli2009fifth} and WNLI~\cite{levesque2012winograd}. The metrics for these tasks can be found in the GLUE paper~\cite{wang2018glue}.

\subsection{TinyBERT Settings}\label{subsec:tinybert_setup}
We instantiate a tiny student model (the number of layers $M$=4, the hidden size $d'$=312, the feed-forward/filter size $d'_i$=1200 and the head number $h$=12) that has a total of 14.5M parameters. This model is referred to as TinyBERT$_{4}$. The original BERT$_{\rm BASE}$ ($N$=12, $d$=768, $d_i$=3072 and $h$=12) is used as the teacher model that contains 109M parameters. We use $g(m)=3\times m$ as the layer mapping function, so TinyBERT$_{4}$ learns from every 3 layers of BERT$_{\rm BASE}$. The learning weight $\lambda$ of each layer is set to 1. Besides, for a direct comparisons with baselines, we also instantiate a TinyBERT$_{6}$~($M$=6, $d'$=768, $d'_i$=3072 and $h$=12) with the same architecture as BERT$_{6}$-PKD~\cite{sun2019patient} and DistilBERT$_{6}$~\cite{sanh2019distilbert}.

TinyBERT learning includes the general distillation and the task-specific distillation. For the general distillation, we set  the maximum sequence length to 128 and use English Wikipedia (2,500M words) as the text corpus and perform the {\it intermediate layer distillation} for 3 epochs with the supervision from a pre-trained BERT$_{\rm BASE}$ and keep other hyper-parameters the same as BERT pre-training~\cite{devlin2019bert}. For the task-specific distillation, under the supervision of a fine-tuned BERT, we firstly perform {\it intermediate layer distillation} on the augmented data for 20 epochs\footnote{For large datasets MNLI, QQP, and QNLI, we only perform 10 epochs of the {\it intermediate layer distillation}, and for the challenging task CoLA, we perform 50 epochs at this step.} with batch size 32 and learning rate 5e-5, and then perform {\it prediction layer distillation} on the augmented data~\footnote{For regression task STS-B, the original train set is better.} for 3 epochs with choosing the batch size from \{16, 32\} and learning rate from \{1e-5, 2e-5, 3e-5\} on dev set. At task-specific distillation, the maximum sequence length is set to 64 for single-sentence tasks, and 128 for sequence pair tasks. 

\subsection{Baselines}\label{baselines}
We compare TinyBERT with BERT$_{\rm TINY}$, BERT$_{\rm SMALL}$\footnote{\url{https://github.com/google-research/bert}}~\cite{turc2019well} and several state-of-the-art KD baselines including BERT-PKD~\cite{sun2019patient}, PD~\cite{turc2019well}, DistilBERT~\cite{sanh2019distilbert} and MobileBERT~\cite{sun2020mobilebert}. BERT$_{\rm TINY}$ means directly pretraining a small BERT, which has the same model architecture as TinyBERT$_{4}$. When training BERT$_{\rm TINY}$, we follow the same learning strategy as described in the original BERT~\cite{devlin2019bert}. To make a fair comparison, we use the released code to train a 4-layer BERT$_{4}$-PKD\footnote{\url{https://github.com/intersun/PKD-for-BERT-Model-Compression}} and a 4-layer DistilBERT$_{4}$\footnote{\url{https://github.com/huggingface/transformers/tree/master/examples/distillation}} and fine-tuning these 4-layer baselines with suggested hyper-paramters. For 6-layer baselines, we use the reported numbers or evaluate the results on the test set of GLUE with released models.


\subsection{Experimental Results on GLUE}
We submitted our model predictions to the official GLUE evaluation server to obtain results on the test set\footnote{\url{https://gluebenchmark.com}}, as summarized in Table~\ref{tab:glue_main}.



The experiment results from the 4-layer student models demonstrate that: 1) There is a large performance gap between BERT$_{\rm TINY}$ (or BERT$_{\rm SMALL}$) and BERT$_{\rm BASE}$ due to the dramatic reduction in model size. 2) TinyBERT$_{4}$ is consistently better than BERT$_{\rm TINY}$ on all the GLUE tasks and obtains a large improvement of 6.8\% on average. This indicates that the proposed KD learning framework can effectively improve the performances of small models on a variety of downstream tasks. 3) TinyBERT$_{4}$ significantly outperforms the 4-layer state-of-the-art KD baselines (i.e., BERT$_{4}$-PKD and DistilBERT$_{4}$) by a margin of at least 4.4\%, with $\sim$28\% parameters and 3.1x inference speedup. 4) Compared with the teacher BERT$_{\rm BASE}$, TinyBERT$_{4}$ is 7.5x smaller and 9.4x faster in the model efficiency, while maintaining competitive performances. 5) For the challenging CoLA dataset~(the task of predicting linguistic acceptability judgments), all the 4-layer distilled models have big performance gaps compared to the teacher model, while TinyBERT$_{4}$ achieves a significant improvement over the 4-layer baselines. 6) We also compare TinyBERT with the 24-layer MobileBERT$_{\rm TINY}$, which is distilled from 24-layer IB-BERT$_{\rm LARGE}$. The results show that TinyBERT$_4$ achieves the same average score as the 24-layer model with only 38.7\% FLOPs. 7) When we increase the capacity of our model to TinyBERT$_{6}$, its performance can be further elevated and outperforms the baselines of the same architecture by a margin of 2.6\% on average and achieves comparable results with the teacher. 8) Compared with the other two-stage baseline PD, which first pre-trains a small BERT, then performs distillation on a specific task with this small model, TinyBERT initialize the student in task-specific stage via general distillation. We analyze these two initialization methods in Appendix~\ref{apx:bert_small}.

In addition, BERT-PKD and DistilBERT initialize their student models with some layers of a pre-trained BERT, which makes the student models have to keep the same size settings of Transformer layer (or embedding layer) as their teacher. In our two-stage distillation framework, TinyBERT is initialized through general distillation, making it more flexible in choosing model configuration.

\noindent{\bf More Comparisons.} We demonstrate the effectiveness of TinyBERT by including more baselines such as Poor Man’s BERT~\cite{sajjad2020poor}, BERT-of-Theseus~\cite{xu2020bert} and MiniLM~\cite{wang2020minilm}, some of which only report results on the GLUE dev set. In addition, we evaluate TinyBERT on SQuAD v1.1 and v2.0. Due to the space limit, we present our results in the Appendix~\ref{apx:more_compar} and \ref{apx:qa_tasks}.

\subsection{Ablation Studies}
In this section, we conduct ablation studies to investigate the contributions of : a) different procedures of the proposed two-stage TinyBERT learning framework in Figure~\ref{figure:tinybert_learning}, and b) different distillation objectives in Equation~\ref{eq:model_loss}.

\subsubsection{Effects of Learning Procedure} The proposed two-stage TinyBERT learning framework consists of three key procedures: GD~(General Distillation), TD~(Task-specific Distillation) and DA~(Data Augmentation). The  performances of removing each individual learning procedure are analyzed and presented in Table~\ref{tab:learning_procedures}. The results indicate that all of the three procedures are crucial for the proposed method. The TD and DA has comparable effects in all the four tasks. We note that the task-specific procedures (TD and DA) are more helpful than the pre-training procedure (GD) on all of the tasks. Another interesting observation is that GD contribute more on CoLA than on MNLI and MRPC. We conjecture that the ability of linguistic generalization~\cite{warstadt2019neural} learned by GD plays an important role in the task of linguistic acceptability judgments.

\begin{table}
\begin{center}
\scalebox{0.75}{
\begin{tabular}{l|cccc|c}
System             	&  MNLI-m    	& MNLI-mm		& MRPC     & CoLA 	& { Avg} \\ 
\hline
TinyBERT$_{4}$      	& 82.8		& 82.9		& 85.8     & 50.8    & 75.6   		\\
\hline
w/o GD             	& 82.5 		& 82.6        	& 84.1     & 40.8    & 72.5     	\\
w/o TD           	& 80.6		& 81.2		& 83.8     & 28.5    & 68.5    		\\
w/o  DA       		& 80.5		& 81.0		& 82.4     & 29.8    & 68.4     	\\
\end{tabular}}
\caption{Ablation studies of different procedures~(i.e., TD, GD, and DA) of the two-stage learning framework. The variants are validated on the dev set.}
\label{tab:learning_procedures}
\vspace{-0.05in}
\end{center} 
\end{table}

\begin{table}
\begin{center} 
\scalebox{0.68}{
\begin{tabular}{l|cccc|c}
	System         &   MNLI-m   &  MNLI-mm   &    MRPC    &    CoLA    &   { Avg}   \\ \hline
	TinyBERT$_{4}$ &    82.8    &    82.9    &    85.8    & 50.8       & 75.6    \\ \hline
	w/o Embd        &    82.3    &    82.3    &    85.0    &    46.7    &    74.1    \\
	w/o Pred        &    80.5    &    81.0    &    84.3    &    48.2    &    73.5    \\
	w/o Trm         &    71.7    &    72.3    &    70.1    &    11.2    &    56.3    \\
	~~~~~~w/o Attn  & ~~~~~~79.9 & ~~~~~~80.7 & ~~~~~~82.3 & ~~~~~~41.1 & ~~~~~~71.0 \\
	~~~~~~w/o Hidn  & ~~~~~~81.7 & ~~~~~~82.1 & ~~~~~~84.1 & ~~~~~~43.7 & ~~~~~~72.9
\end{tabular}}
\caption{Ablation studies of different distillation objectives in the TinyBERT learning. The variants are validated on the dev set.}
\label{exp:distill_ablation}   
\vspace{-0.15in}
\end{center} 
\end{table}

\subsubsection{Effects of Distillation Objective} We investigate the effects of distillation objectives on the TinyBERT learning. Several baselines are proposed including the learning without the Transformer-layer distillation (w/o Trm), the embedding-layer distillation (w/o Emb) or the prediction-layer distillation (w/o Pred)\footnote{The prediction-layer distillation performs soft cross-entropy as Equation~\ref{eq:pred_loss} on the augmented data. ``w/o Pred'' means performing standard cross-entropy against the ground-truth of original train set.} respectively. The results are illustrated in Table~\ref{exp:distill_ablation} and show that all the proposed distillation objectives are useful. 
The performance w/o Trm\footnote{Under ``w/o Trm" setting, we actually 1) conduct embedding-layer distillation at the pre-training stage; 2) perform embedding-layer and prediction-layer distillation at fine-tuning stage.} drops significantly from 75.6 to 56.3. The reason for the significant drop lies in the initialization of student model. At the pre-training stage, obtaining a good initialization is crucial for the distillation of transformer-based models, while there is no supervision signal from upper layers to update the parameters of transformer layers at this stage under the w/o Trm setting. Furthermore, we study the contributions of attention (Attn) and hidden states (Hidn) in the Transformer-layer distillation. We can find the attention based distillation has a greater impact than hidden states based distillation. Meanwhile, these two kinds of knowledge distillation are complementary to each other, which makes them the most important distillation techniques for Transformer-based model in our experiments. 


\subsection{Effects of Mapping Function}
\label{exp:distill_layers}
We also investigate the effects of different mapping functions $n=g(m)$ on the TinyBERT learning. Our original TinyBERT as described in section~\ref{subsec:tinybert_setup} uses the uniform strategy, and we compare with two typical baselines including top-strategy $\left(g(m)=m+N-M; 0<m \leq M \right)$ and bottom-strategy $\left(g(m)=m; 0<m \leq M \right)$. 

The comparison results are presented in Table~\ref{table:distill_layers}. We find that the top-strategy performs better than the bottom-strategy on MNLI, while being worse on MRPC and CoLA, which confirms the observations that different tasks depend on the knowledge from different BERT layers. The uniform strategy covers the knowledge from bottom to top layers of BERT$_{\rm BASE}$, and it achieves better performances than the other two baselines in all the tasks. Adaptively choosing layers for a specific task is a challenging problem and we leave it as future work.

\begin{table}
\begin{center}
\scalebox{0.8}{
 \begin{tabular}{l|cccc|c}
   System & MNLI-m & MNLI-mm & MRPC & CoLA 	& {Avg}    \\
\hline
Uniform     &  82.8		&  82.9		&  85.8	& 50.8    & 75.6 \\
\hline
Top         & 81.7	 	& 82.3		& 83.6	& 35.9	& 70.9 \\
Bottom   	  & 80.6		& 81.3		& 84.6	& 38.5	& 71.3 \\
\end{tabular}}
\caption{Results (dev) of different mapping strategies for TinyBERT$_{4}$.}
\label{table:distill_layers}  
\vspace{-0.15in} 
\end{center} 
\end{table}

\section{Related Work}\label{sec:related}

\noindent{\bf Pre-trained Language Models Compression}
Generally, pre-trained language models~(PLMs) can be compressed by low-rank approximation~\cite{ma2019tensorized,Lan2020ALBERT}, weight sharing~\cite{dehghani2018universal,Lan2020ALBERT}, knowledge distillation~\cite{tang2019distilling,sanh2019distilbert,turc2019well,sun2020mobilebert,liu2020fastbert,wang2020minilm}, pruning~\cite{cui2019fine,mccarley2019pruning,Fan2020Reducing,Elbayad2020Depth-Adaptive,gordon2020compressing,hou2020dynabert} or quantization~\cite{shen2019q,zafrir2019q8bert}. In this paper, our focus is on knowledge distillation.

\noindent{\bf Knowledge Distillation for PLMs}
There have been some works trying to distill pre-trained language models~(PLMs) into smaller models. BiLSTM$_{\tiny \hbox{SOFT}}$~\cite{tang2019distilling} distills task-specific knowledge from BERT into a single-layer BiLSTM. BERT-PKD~\cite{sun2019patient} extracts knowledges not only from the last layer of the teacher, but also from intermediate layers at fine-tuning stage. DistilBERT~\cite{sanh2019distilbert} performs distillation at pre-training stage on large-scale corpus. Concurrent works, MobileBERT~\cite{sun2020mobilebert} distills a BERT$_{\rm LARGE}$ augmented with bottleneck structures into a 24-layer slimmed version by progressive knowledge transfer at pre-training stage. MiniLM~\cite{wang2020minilm} conducts deep self-attention distillation also at pre-training stage. By contrast, we propose a new {\it two-stage learning} framework to distill knowledge from BERT at both pre-training and fine-tuning stages by a novel transformer distillation method. 

\noindent{\bf Pretraining Lite PLMs}
Other related works aim at directly pretraining lite PLMs. \citet{turc2019well} pre-trained 24 miniature BERT models and show that pre-training remains important in the context of smaller architectures, and fine-tuning pre-trained compact models can be competitive. ALBERT~\cite{Lan2020ALBERT} incorporates embedding factorization and cross-layer parameter sharing to reduce model parameters. Since ALBERT does not reduce hidden size or layers of transformer block, it still has large amount of computations. Another concurrent work, ELECTRA~\cite{clark2020electra} proposes a sample-efficient task called replaced token detection to accelerate pre-training, and it also presents a 12-layer ELECTRA$_{\rm small}$ that has comparable performance with TinyBERT$_{4}$. Different from these small PLMs, TinyBERT$_{4}$ is a 4-layer model which can achieve more speedup.


\section{Conclusion and Future Work}
In this paper, we introduced a new method for Transformer-based distillation, and further proposed a two-stage framework for TinyBERT. Extensive experiments show that TinyBERT achieves competitive performances meanwhile significantly reducing the model size and inference time of BERT$_{\rm BASE}$, which provides an effective way to deploy BERT-based NLP models on edge devices. 
In future work, we would study how to effectively transfer the knowledge from wider and deeper teachers (e.g., BERT$_{\rm LARGE}$) to student TinyBERT. Combining distillation with quantization/pruning would be another promising direction to further compress the pre-trained language models.
\label{sec:con}

\section*{Acknowledgements}
This work is supported in part by NSFC NO.61832020, No.61821003, 61772216, National Science and Technology Major Project No.2017ZX01032-101, Fundamental Research Funds for the Central Universities.

\bibliography{emnlp2020}
\bibliographystyle{acl_natbib}

\appendix
\section*{Appendix}
\label{sec:appx}

\section{More Comparisons on GLUE}
\label{apx:more_compar}

\begin{table*}[ht]
	\begin{center}
		\scalebox{0.9}{
			\begin{tabular}{lcccccclcc}
				\hline
				System                       &  CoLA  & MNLI-m & MNLI-mm &   MRPC    &  QNLI  &    QQP    & RTE    & SST-2 &   STS-B   \\
				                             & (8.5k) & (393k) & (393k)  &  (3.7k)   & (105k) &  (364k)   & (2.5k) & (67k) &  (5.7k)   \\
				                             &  Mcc   &  Acc   &   Acc   &  F1/Acc   &  Acc   &  F1/Acc   & Acc    &  Acc  & Pear/Spea \\ \hline
				\multicolumn{10}{l}{\it Same Student Architecture ($M$=6;$d'$=768;$d'_i$=3072)}                                        \\ \hline
				DistilBERT$_{6}$             &  51.3  &  82.2  &  -      & 87.5/-    &  89.2  & -/88.5    & 59.9   & 92.7  & -/86.9    \\
				Poor Man’s BERT$_{6}$        &  -     &  81.1  &  -      & -/80.2    &  87.6  & -/90.4    & 65.0   & 90.3  & -/88.5    \\
				BERT-of-Theseus              &  51.1  &  82.3  &  -      & 89.0/-    &  89.5  & -/89.6    & 68.2   & 91.5  & -/88.7    \\
				
				MiniLM$_{6}$                 &  49.2  &  84.0  &    -    &  88.4/-   &  91.0  &  -/91.0   & 71.5   & 92.0  &     -     \\
				TinyBERT$_{6}$               &  \textbf{54.0}  &  \textbf{84.5}  &  \textbf{84.5}   & \textbf{90.6/86.3} &  \textbf{91.1}  & \textbf{88.0/91.1} & \textbf{73.4}   & \textbf{93.0}  & \textbf{90.1/89.6} \\ \hline
			\end{tabular}}
		\caption{Comparisons between TinyBERT with other baselines on the dev set of GLUE tasks. Mcc refers to Matthews correlation and Pear/Spea refer to Pearson/Spearman.}
		\label{tab:compare_dev}
	\end{center}
\end{table*}

Since some prior works on BERT compression only evaluate their models on the GLUE dev set, for an easy and direct comparison, we here compare our TinyBERT$_6$ with the reported results from these prior works. All the compared methods have the same model architecture as TinyBERT$_6$ (i.e. $M$=6, $d'$=768, $d'_i$=3072). 

The direct comparison results are shown in Table~\ref{tab:compare_dev}. We can see the TinyBERT$_6$ outperforms all the baselines under the same settings of architecture and evaluation methods. The effectiveness of TinyBERT is further confirmed. 


\section{Results on SQuAD v1.1 and v2.0}
\label{apx:qa_tasks}
We also demonstrate the effectiveness of TinyBERT on the question answering (QA) tasks: SQuAD v1.1~\citep{rajpurkar2016squad} and SQuAD v2.0~\citep{rajpurkar2018know}. Following the learning procedure in the previous work~\citep{devlin2019bert}, we treat these two tasks as the problem of sequence labeling which predicts the possibility of each token as the start or end of answer span. 
One small difference from the GLUE tasks is that we perform the prediction-layer distillation on the original training dataset instead of the augmented dataset, which can bring better performances. 


The results show that TinyBERT consistently outperforms both the 4-layer and 6-layer baselines, which indicates that the proposed framework also works for the tasks of token-level labeling. Compared with sequence-level GLUE tasks, the question answering tasks depend on more subtle knowledge to infer the correct answer, which increases the difficulty of knowledge distillation. We leave how to build a better QA-TinyBERT as future work.
\begin{table}
\begin{center}
\scalebox{0.85}{
 \begin{tabular}{l|cccc}
 	\hline
 	System                      & \multicolumn{2}{c}{\textbf{SQuAD 1.1}} & \multicolumn{2}{c}{\textbf{SQuAD 2.0}} \\
 	                            &      EM       &           F1           &      EM       &           F1           \\ \hline
 	BERT$_{\rm BASE}$ (Teacher) &     80.7      &          88.4          &     74.5      &          77.7          \\ \hline
 	\multicolumn{5}{l}{\it 4-layer student models}                                                                \\ \hline
 	BERT$_{4}$-PKD              &     70.1      &          79.5          &     60.8      &          64.6          \\
 	DistilBERT$_{4}$            &     71.8      &          81.2          &     60.6      &          64.1          \\
 	MiniLM$_{4}$         &       -       &           -            &       -       &          69.7          \\
 	TinyBERT$_{4}$              & \textbf{72.7} &     \textbf{82.1}      & \textbf{68.2} &     \textbf{71.8}      \\ \hline
 	\multicolumn{5}{l}{\it 6-layer student models}                                                                \\ \hline
 	BERT$_{6}$-PKD              &     77.1      &          85.3          &     66.3      &          69.8          \\
 	DistilBERT$_{6}$            &     78.1      &          86.2          &     66.0      &          69.5          \\
 	MiniLM$_{6}$                &       -       &           -            &       -       &          76.4          \\
 	TinyBERT$_{6}$              & \textbf{79.7} &     \textbf{87.5}      & \textbf{74.7} &     \textbf{77.7}      \\ \hline
 \end{tabular}}
   \caption{Results (dev) of baselines and TinyBERT on question answering tasks. The architecture of MiniLM$_{4}$ is ($M$=4, $d$=384, $d_i$=1536) which is wider than TinyBERT$_{4}$, and the architecture of MiniLM$_{6}$ is the same as TinyBERT$_6$($M$=6, $d$=768, $d_i$=3072)}
   \label{exp:squad}
\end{center}
\end{table}

\section{Initializing TinyBERT with BERT$_{\rm TINY}$}
\label{apx:bert_small}

In the proposed two-stage learning framework, to make TinyBERT effectively work for different downstream tasks, we propose the General Distillation~(GD) to capture the general domain knowledge, through which the TinyBERT learns the knowledge from intermediate layers of teacher BERT at the pre-training stage. After that, a general TinyBERT is obtained and used as the initialization of student model for Task-specific Distillation~(TD) on downstream tasks. 


\begin{table}[t]
	\begin{center}
		\scalebox{0.66}{
			\begin{tabular}{@{}l|cccc|c}
				\hline
				System                 & MNLI-m & MNLI-mm &  MRPC  &  CoLA  & { Avg} \\
				& (392k) & (392k)  & (3.5k) & (8.5k) &        \\ \hline
				BERT$_{\rm TINY}$      &  75.9  &  76.9   &  83.2  &  19.5  &  63.9  \\
				BERT$_{\rm TINY}$(+TD) &  79.2  &  79.7   &  82.9  &  12.4  &  63.6  \\ \hline
				TinyBERT~(GD)          &  76.6  &  77.2   &  82.0  &  8.7   &  61.1  \\
				TinyBERT~(GD+TD)       &  80.5  &  81.0   &  82.4  &  29.8  &  68.4  \\ \hline
		\end{tabular}}
		\caption{Results of different methods at pre-training stage. TD and GD refers to Task-specific Distillation (without data augmentation) and General Distillation, respectively. The results are evaluated on dev set.}
		\label{tab:general_distill}
	\end{center}
\end{table}

In our preliminary experiments, we have also tried to initialize TinyBERT with the directly pre-trained BERT$_{\rm TINY}$, and then conduct the TD on downstream tasks. We denote this compression method as BERT$_{\rm TINY}$(+TD). The results in Table~\ref{tab:general_distill} show that BERT$_{\rm TINY}$(+TD) performs even worse than  BERT$_{\rm TINY}$ on MRPC and CoLA tasks. We conjecture that if without imitating the BERT$_{\rm BASE}$'s behaviors at the pre-training stage, BERT$_{\rm TINY}$ will derive mismatched distributions in intermediate representations (e.g., attention matrices and hidden states) with the BERT$_{\rm BASE}$ model. The following task-specific distillation under the supervision of fine-tuned BERT$_{\rm BASE}$ will further disturb the learned distribution/knowledge of BERT$_{\rm TINY}$, finally leading to poor performances on some less-data tasks. For the intensive-data task~(e.g. MNLI), TD has enough training data to make BERT$_{\rm TINY}$ acquire the task-specific knowledge very well, although the pre-trained distributions have already been disturbed.

 From the results of Table~\ref{tab:general_distill}, we find that GD can effectively transfer the knowledge from the teacher BERT to the student TinyBERT and achieve comparable results with BERT$_{\rm TINY}$~(61.1 vs. 63.9), even without performing the MLM and NSP tasks. Furthermore, the task-specific distillation boosts the performances of TinyBERT by continuing on learning the task-specific knowledge from fine-tuned teacher BERT$_{\rm BASE}$.

\section{GLUE Details}
\label{apx:glue}

The GLUE datasets are described as follows:

\noindent\textbf{MNLI.} Multi-Genre Natural Language Inference is a large-scale, crowd-sourced entailment classification task \citep{williams2018broad}. Given a pair of $\langle premise, hypothesis \rangle$, the goal is to predict whether the $hypothesis$ is an entailment, contradiction, or neutral with respect to the $premise$.

\noindent\textbf{QQP.} Quora Question Pairs is a collection of question pairs from the website Quora. The task is to determine whether two questions are semantically equivalent \citep{chen2018quora}.

\noindent\textbf{QNLI.} Question Natural Language Inference is a version of the Stanford Question Answering Dataset which has been converted to a binary sentence pair classification task by \citet{wang2018glue}. Given a pair $\langle question, context \rangle$. The task is to determine whether the $context$ contains the answer to the $question$.

\noindent\textbf{SST-2.} The Stanford Sentiment Treebank is a binary single-sentence classification task, where the goal is to predict the sentiment of movie reviews~\cite{socher2013recursive}.

\noindent\textbf{CoLA.} The Corpus of Linguistic Acceptability is a task to predict whether an English sentence is a grammatically correct one~\citep{warstadt2019neural}.

\noindent\textbf{STS-B.} The Semantic Textual Similarity Benchmark is a collection of sentence pairs drawn from news headlines and many other domains \citep{cer2017semeval}. The task aims to evaluate how similar two pieces of texts are by a score from 1 to 5.

\noindent\textbf{MRPC.} Microsoft Research Paraphrase Corpus is a paraphrase identification dataset where systems aim to identify if two sentences are paraphrases of each other~\cite{dolan2005automatically}.

\noindent\textbf{RTE.} Recognizing Textual Entailment is a binary entailment task with a small training dataset~\citep{bentivogli2009fifth}.

\end{document}